\documentclass[preprint,review,10pt]{elsarticle}

\usepackage[lined,  boxed,  linesnumbered,  ruled]{algorithm2e}
\usepackage{lineno,hyperref}
\usepackage{amssymb, amsfonts, amsmath, graphicx, epstopdf, subfig, bbding, lipsum, multirow, makecell, picinpar, enumerate}
\usepackage[table]{xcolor}
\usepackage{colortbl}

\journal{Pattern Recognition}

\begin{document}
\begin{frontmatter}



\title{GAN for Vision, KG for Relation:\\ a Two-stage Deep Network for Zero-shot Action Recognition}


\author[label1]{Bin Sun}
\author[label1]{Dehui Kong}
\author[label1]{Shaofan Wang\corref{cor1}}\ead{wangshaofan@bjut.edu.cn} \cortext[cor1]{Corresponding author.}
\author[label1]{Jinghua Li}
\author[label1]{Baocai Yin}
\author[label1]{Xiaonan Luo}
\address[label1]{Beijing Key Laboratory of Multimedia and Intelligent Software Technology, Faculty of Information Technology, \\ Beijing University of Technology, Beijing 100124, China}

\begin{abstract}
Zero-shot action recognition can recognize samples of unseen classes that are unavailable in training by exploring common latent semantic representation in samples. However, most methods neglected the connotative relation and extensional relation between the action classes, which leads to the poor generalization ability of the zero-shot learning. Furthermore, the learned classifier incline to predict the samples of seen class, which leads to poor classification performance.
To solve the above problems, {we propose a two-stage deep neural network for zero-shot action recognition, which consists of a feature generation sub-network serving as the sampling stage and a graph attention sub-network serving as the classification stage.
In the sampling stage, we utilize a generative adversarial networks (GAN) trained by action features and word vectors of seen classes to synthesize the action features of unseen classes, which can balance the training sample data of seen classes and unseen classes.
In the classification stage, we construct a knowledge graph (KG) based on the relationship between word vectors of action classes and related objects, and propose a graph convolution network (GCN) based on attention mechanism, which dynamically updates the relationship between action classes and objects, and enhances the generalization ability of zero-shot learning.
In both stages, we all use word vectors as bridges for feature generation and classifier generalization from seen classes to unseen classes.}
We compare our method with state-of-the-art methods on \texttt{UCF101}  and \texttt{HMDB51} datasets.
Experimental results show that our proposed method improves the classification performance of the trained classifier and achieves higher accuracy.
\end{abstract}

\begin{keyword}
Action recognition \sep Zero-shot learning  \sep Generative adversarial networks \sep Graph convolution network

\end{keyword}

\end{frontmatter}





\section{Introduction}
Action recognition is an important research issue in the fields of machine learning and computer vision, and has wide applications in human-computer interaction, video monitoring, motion retrieval and sports video analysis \cite{ji2011actor,wang2013learning,
wang2019order,li2020spatio,wang2020deep,wang2017g2denet}. With the rapid development of internet technology and emerging social media, monocular or multiview videos provide fruitful cues for action recognition\cite{liu2016simple,yao2017learn,sun2019effective}. Hence, the increase of both the complexity of human actions and the number of video classes is also inevitable. On the other hand, {annotating massive video data is pervasive and important for action recognition, but it is a tedious and inaccurate task}. Since it is not only a time-consuming and expensive operation, but also easy to be influenced by subjective {judgments} of experts. At the same time, due to the limitation of extensibility of data classes, traditional action recognition methods are unsuitable for recognizing data from unseen classes and cannot support the realization of automatic annotation. Therefore, how to obtain potential information from labeled videos so as to effectively annotate unseen videos has become an urgent problem.

Zero-shot action recognition \cite{long2017zero,long2017learning,tian2020aligned} provides an effective means to settle this problem, which can recognize data of unseen classes (the classes without training samples), and has attracted extensive attention recently. The aim of zero-shot learning (ZSL) is to explore common latent semantic representation, and {produce} a trained model that can generalize to unseen classes. Existing zero-shot action recognition methods can be divided into two categories.
Manually-defined attribute based methods utilize manually defined attributes for classification, which are easy to understand and implement, and only use the relationship between actions and attributes to distinguish new action classes. However, human subjectivity and the lack of domain knowledge make it difficult to identify a set of attributes that can describe all actions. In addition, although attributes can be viewed as data-driven learning, their semantic meaning may be unseen or inappropriate. Therefore, it is difficult to generalize attribute-based methods to  large-scale scenarios.
Word embedding based methods utilize  semantic representation of action class names (e.g., word vectors) to model the relationship between actions and actions in  semantic {spaces}. The word vectors used in these methods are acquired by natural language processing using massive amounts of text information. Therefore, these methods can overcome the limitation of attribute based methods. However, these methods can only express the relationship between actions implicitly in {word vector spaces} and hardly benefit from the other information of videos resulting in poor classification performance.

In summary, the above research work on zero-shot action recognition neglect the connotative relation and extensional relation between {action classes}, which leads to a poor generalization ability of ZSL. In fact, humans can use empirically learned semantic knowledge to extend their ability to recognize large-scale concepts by virtue of the association between connotation and extension extension. Thus, using structured knowledge information to build relationships of concepts (e.g. actions and attributes) can transfer learned knowledge from seen classes to unseen classes. Recently, {Graph Convolutional Networks (GCN) based methods~\cite{kipf2016semi,gao2019know} applied knowledge graph (KG) to ZSL and achieved promising results.} However, the adjacency matrix constructed by these methods remains unchanged after initial setting, which makes it unable to describe the changing relationships of nodes in the graph adaptively, resulting in incomplete knowledge transfer.
In addition, most of ZSL methods cannot use the samples of unseen classes in training, which makes the training classifier more inclined to predict the samples of seen classes. {Recently, Generative Adversarial Networks (GAN) based methods~\cite{felix2018multi,mandal2019out} synthesized the features of unseen classes, which are used to train the classifier. The performance of these methods can effectively improved, which shows that providing the samples of unseen classes during training can make the learned classifier better adapt to the classification requirements of unseen classes. }

\begin{figure*}
\centering
\includegraphics[scale=0.7]{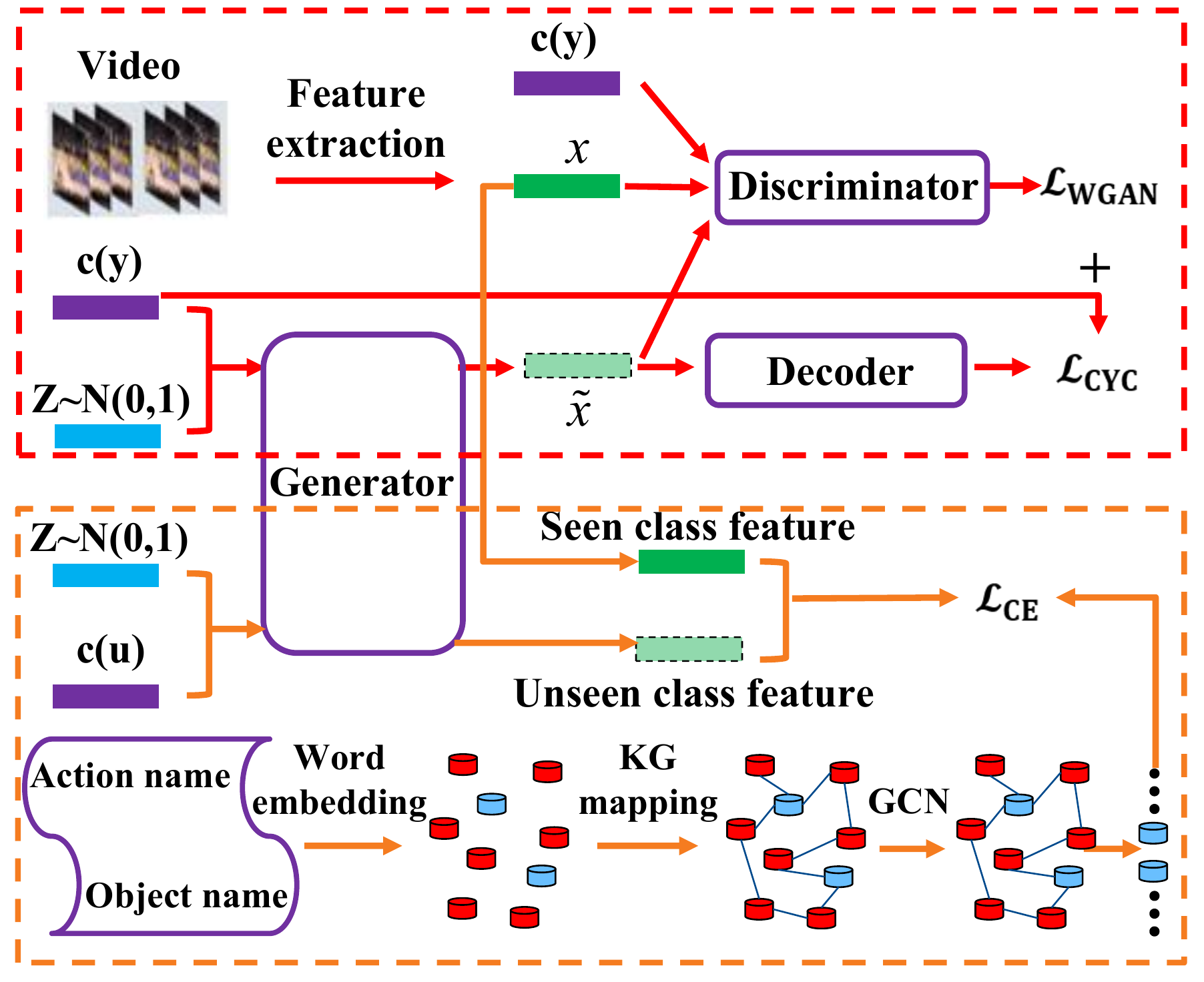}
\caption{Illustration of FGGA. {FGGA consists of two stages: sampling stage and classification stage. The red arrows indicate the sampling stage, which is the training process of feature generation network, The orange arrows indicate the classification stage, which is the training process of graph attention convolution network. Red cylinder and blue cylinder represent the object node and the class node, respectively.}}
\label{framework}
\end{figure*}

{It can be seen from the above that GCN based methods and GAN based methods have different perspectives of analysis and thinking. GAN based zero-shot learning used the adversarial relationship between sample features to realize the generalization of the feature generation ability of seen class to unseen class. Knowledge graph based zero-shot learning used the extrinsic correlation between classes and attributes to realize the generalization of classification ability. In general, GAN based methods are analyzed from sample level, and GCN based methods are analyzed from the classifier level. For both of these methods, the word vector is a bridge from the seen classes to the unseen classes for feature generation or classifier generalization. On the whole, these two types of methods are complementary with each other and the performance will be better by exploring them simultaneously.
Therefore, we consider comprehensively from two aspects, and propose a new zero-shot action recognition, i.e., joint Feature Generation network and Graph Attention network (FGGA) for ZSL. The main idea comes from \cite{Feichtenhofer2016conv,hao2019spatiotemporal} which indicated that simultaneously exploiting information from different perspectives can complementary with each other for action recognition than single exploitation.}
On one hand, FGGA utlizes GAN to synthesize the features of unseen classes, which can reduce the imbalance between training samples of seen classes and unseen classes. On the other hand, {FGGA constructs a KG based on the relationship between action class and related objects, and a graph convolution network} based on attention mechanism. In fact, graph convolution network based on attention mechanism can effectively realize the dynamic expression of the relation between action classes and objects, which reflects the influence of knowledge update on model learning.
Experimental results show that FGGA improves the classification performance of the trained classifier and can achieve higher accuracy. A flowchart of FGGA is illustrated in Figure~\ref{framework}. {FGGA consists of two stages: sampling stage and classification stage. In the sampling stage, FGGA uses feature generation network to train generator, which is used to synthesize the features of unseen classes. In the second stage, FGGA uses graph attention network to train classifier, which is used to expand the generalization ability of the classifier.}

The main contributions of our work are summarized as follows.
\begin{enumerate}[$\bullet$]
\item {We propose a two-stage network for zero-shot action recognition: FGGA, which joints feature generation network
and graph attention network, makes a comprehensive analysis from sample level and classifier level.}
\item  {FGGA adopts the conditional Wasserstein GAN with additional loss terms to train generator, which can transform the representation space of actions from the word vector space to the visual feature space, and the synthesized features of unseen class can be straightforwardly fed to typical classifiers.}
\item  FGGA integrates an attention mechanism with GCN, and expresses the relationship between action class and related objects dynamically.
\end{enumerate}

The remainder of this paper is organized as follows. Section~\ref{Related Work} introduces the related work. Section~\ref{Methodology} elaborates FGGA. Section~\ref{Experiments} conducts comparative experiments and evaluates the performance of FGGA. Section~\ref{Conclusion} gives the conclusion.

\section{Related Work}\label{Related Work}
In this section, we briefly review the work related to our proposed method. We review ZSL methods, which can be roughly grouped into two categories: attribute based methods and word embedding based methods.
\subsection{Attribute based methods}
Early research work on action recognition mainly considered a set of manually defined attributes to describe the spatio-temporal evolution of actions. Liu~{\it et~al}.~\cite{liu2011recognizing} proposed a potential SVM model in which potential variables determine the importance of each attribute of each class. Lampert~{\it et~al}.~\cite{lampert2013attribute} proposed a direct and indirect attribute prediction model, which used attribute-class and class-class correlation to predict unseen samples. Akata~{\it et~al}.~\cite{akata2013label} regarded the classification problem based on attributes as a label embedding problem, and each class was embedded in the space of attribute vectors. This method measured the matching degree of image and label embedding by constructing matching function. Gan~{\it et~al}.~\cite{gan2016learning} studied how to accurately and robustly detect attributes in images or video, and the learned high-quality attribute detector can be generalized to different classes. The advantage of the attribute-based approach is that each attribute describes the shared features between classes, and it can transfer knowledge between classes well. In addition, attributes can also describe the details of the class, which can be used to predict the unseen class. However, the attribute-based method has several drawbacks: Firstly, each action is a complex combination of various human motions and human-object interactions. Human subjective factors and the lack of domain knowledge make it very difficult to determine a set of attributes used to describe all actions.
Secondly, the relationship between classes and attributes needs to be defined manually, whenever a new class is created, the attribute of class and the corresponding relationship between attribute and class need to be changed again. Therefore,  attribute-based methods are not applicable to ZSL problems of massive classes.
Thirdly, attributes can be viewed as data-driven learning, but their semantic meaning may be unknown or inappropriate.

\subsection{Word embedding based methods}
Since only action class names are needed to construct label embedding, word embedding based methods are gradually favored.
Xu~{\it et~al}.~\cite{xu2015semantic} first applied word vector space as middle layer embedding to the zero shot learning for action recognition. Subsequently, Xu~{\it et~al}.~\cite{xu2017transductive} proposed manifold regularization regression and data enhancement strategies, which significantly enhanced semantic space mapping, and used post-processing strategies such as self-training to further improve accuracy. Alexiou~{\it et~al}.~\cite{alexiou2016exploring} explored broader semantic context information in the text domain (for example, synonyms) expressed as word vectors that enrich the action class. Qin~{\it et~al}.~\cite{qin2017zero} adopted error-correcting output coding to solve the domain shift problem, which took the advantage of  semantics of class layers and internal data structure. However, because of the semantic differences between visual and textual information, individual word vectors are not sufficient to distinguish between classes. Recently, inspired by the strong relationship between objects and action, some methods have achieved good performance by using objects as attributes. Jaine~{\it et~al}.~\cite{jain2015objects2action} constructed a semantic embedding model by considering thousands of object classes. Mettes~{\it et~al}.~\cite{mettes2017spatial} further designed spatial perception object embedding to classify actions. In addition, a few methods explored semantic relationships using inter-class relationships~\cite{gan2015exploring} and pair relationships~\cite{gan2016concepts}. Gan~{\it et~al}.~\cite{gan2016recognizing} used knowledge information to construct an analogy pool based on external ontology. However, these methods are not end-to-end.
Gao~{\it et~al}.~\cite{gao2019know} proposed a ZSL framework based on two-stream graph convolutional networks and knowledge graphs, and established a relationship between objects and classes in an end-to-end manner.
{Oza~{\it et~al}.~\cite{oza2019c2ae} used class conditioned auto-encoders and the training procedure was divided in two sub-tasks, where encoder learns the first task following the closed-set classification training pipeline, and decoder learns the second task by reconstructing conditioned on class identity.}
Mandal~{\it et~al}.~\cite{mandal2019out} proposed an out-of-distribution (OD) detector for detecting features of unseen classes, which could be trained by using real features of seen classes and synthetic features of unseen classes. Compared with {shared semantic embedding spaces} consisting of manually defined class attributes, {word embedding spaces} are composed of the word vectors of all classes. In {these spaces}, each class is described by its word vector. These word vectors are learned by natural language processing techniques using massive amounts of text information. Therefore,  ZSL  {methods} based on word embedding {spaces}  overcome the limitation of  ZSL  {methods} based on attribute {spaces},  {as the latter require manual definitions of} the corresponding relationship between attributes and classes.

{In summary, the state-of-the-art methods mainly use GAN or knowledge graph. However, these methods used GAN and knowledge graph for the problem of zero-shot learning independently. GAN based zero-shot learning~\cite{mandal2019out} considered the problem from the perspective of missing unseen class samples, and synthesized visual representations of the unseen classes. Knowledge graph based zero-shot learning~\cite{gao2019know} considered the problem from the perspective of the unseen classes and its relationship to the seen classes, and built structured knowledge information of unseen classes and seen classes. Generally, GAN based method is analyzed from the perspective of sample, and knowledge graph based method is analyzed from the perspective of class. For both of these methods, the word vector is a bridge from the seen classes to the unseen classes for feature generation or classifier generalization. In this paper, We combine GAN and knowledge graph and propose joint feature generation network and graph attention network for zero-shot
action recognition. We use word vectors as bridges, both the vision features and the classifiers of unseen actions are inferable from seen actions.
Furthermore, the edges of knowledge graph of previous work are constant values, which means the relationships between nodes in the knowledge graph are constant. Recently, incorporation of attention~\cite{velivckovic2018graph,wang2020eca} into Deep learning model has attracted a lot of interests, showing great potential in performance improvement. In this paper, we use graph attention to dynamically expresses the relationship between nodes in the knowledge graph.
Overall, our method enhances the generalization ability and discriminability of the model by dynamically updating the relationship between {action classes and objects} and synthesizing features of unseen classes.}



\section{Methodology}\label{Methodology}
We give the methodology of FGGA in this section. We denote scalars, vectors,  matrices, and sets  by nonbold  letters, bold lowercase letters, bold uppercase letters, and calligraphic uppercase letters respectively. Denote $[\mathbf{A}]_{ij}$ to be the $(i,j)$th element of a matrix $\mathbf{A}$. Denote $\mathbf{I}$ to be the identity matrix whose size is determined in context. Denote $|\cdot|$ to be the cardinality operator.
\subsection{Problem definition}
Let $\mathcal{X}\subseteq \mathbb{R}^{{d_x}}$ be the set of all samples, and let $\mathcal{Y}^s,\mathcal{Y}^u$  be the set of labels of  seen classes, labels of  unseen classes, respectively, where the superscripts $s,u$  denote ``seen", ``unseen", respectively. Let $\mathcal{E}\subseteq {\mathbb{R}^{{d_c}}}$ be the set of word vectors of all classes.  Denote
\begin{equation}
\mathcal{S} = \{ {(\mathbf{x},y,\mathbf{c}(y))|\mathbf{x} \in \mathcal{X},y \in {\mathcal{Y}^s},\mathbf{c}(y) \in \mathcal{E}\} }
\end{equation}
to be the training set for seen classes, where $\mathbf{x}$ represents feature of sample, $y$ represents the class-specific label of seen classes, $\mathbf{c}(y)$ represents the class-specific embedding.
Additionally, we denote
\begin{equation}
{\cal U} = \{ {(u,\mathbf{c}(u))|u \in {\mathcal{Y}^u},\mathbf{c}(u) \in \mathcal{E}\} }
\end{equation}
to be a set for unseen classes, which is available during training, where $u$ represents the class-specific label of unseen classes,  $\mathbf{c}(u) \in {\mathbb{R}^{{d_c}}}$ represents the class-specific embedding, and $\mathcal{Y}^{u} \cap \mathcal{Y}^{s}=\varnothing$.
Furthermore, we have an object set ${\cal O} $, which serves as attributes for describing different action classes.
The goal of ZSL is to learn a classifier
\begin{equation}
f_{\rm ZSL}: \mathcal{X} \rightarrow \mathcal{Y}^{u},\end{equation}
and the goal of generalized zero-shot learning (GZSL) is to learn a classifier
\begin{equation}
f_{\rm GZSL}: \mathcal{X} \rightarrow \mathcal{Y}^{s} \cup \mathcal{Y}^{u}, \end{equation}
where the features of samples of unseen classes $\mathcal{X}^{u}$ are completely unavailable during training.

\subsection{Unseen class feature synthesis based on feature generation network}
Most of the ZSL methods train classifier only by data of seen classes during training, which is prone to be biased towards the samples of seen classes and not conducive to the samples of unseen classes.
To solve this problem, sample imbalance between seen and unseen classes during training needs further exploration. One of the straightforward ideas is to add additional synthesize features of unseen
classes without access to any samples of that class. As we all know, GAN can synthesize images of  unseen
objects through given semantic descriptions. In this paper, the objective of our study is human actions rather than images. Therefore, to apply GAN to our model, we want to synthesize features of samples using GAN.

Given the training data of  {seen classes}, we hope to use the word {vectors} of {unseen classes} to synthesize the features of  unseen classes. Therefore, we use features and word vectors of {seen classes} to learn GAN. GAN consists of {a} generator G and {a} discriminator D, where {the} generator is used to generate ``false" samples and {the} discriminator is used to determine whether samples are real or synthetic. G and D constitute a dynamic ``game process". Under the background of synthesizing video features, the goal of G is to obtain random signals from  prior distribution and synthesize video features close to  real features to deceive the discriminating network D. The goal of D is to accurately distinguish the video features synthesized by G from real video features. Since we need to synthesize the features of {unseen classes}, we use the condition GAN~\cite{mirza2014conditional}, that is, the word vector $\mathbf{c}(y)$ of {seen classes} is added in the modeling of G and D. The conditional generator takes random Gaussian noise and the word vector of seen classes as its input and video feature as its output. Then, the generator can synthesize the video features of {unseen classes} through the word {vectors} of {unseen classes}. The loss function of condition GAN is:
\begin{align}
{\mathcal{L}_{\rm GAN}} = \mathbb{E}[\log D(\mathbf{x},\mathbf{c}(y))]
+ \mathbb{E}[\log (1 - D(\mathbf{\tilde x},\mathbf{c}(y)))]
\end{align}
where $\mathbf{\tilde x} = G(\mathbf{z},\mathbf{c}(y))$ represents the synthetic feature, $\mathbb{E}[\cdot]$ is the expectation, and $D:\mathcal{X} \times \mathcal{E} \to [0,1]$ is a multilayer perceptron with the sigmoid function as the last layer. Although condition GAN can obtain complex data distribution, it is difficult to train. Therefore, we use  Wasserstein Generative Adversarial Network-Gradient Penalty (WGAN-GP, \cite{gulrajani2017improved}) for training, since the algorithm is more stable when training. The loss function of WGAN-GP is:
\begin{align}
{\mathcal{L}_{\rm WGAN}} = \mathbb{E}[D(\mathbf{x},\mathbf{c}(y))] - \mathbb{E}[D(\mathbf{\tilde x}, \mathbf{c}(y))]
- \lambda \mathbb{E}\left[ {{{\left( {{{\left\| {{\nabla _{\mathbf{\widehat{x}}}}D(\mathbf{\widehat{x}}, \mathbf{c}(y))} \right\|}_2} - 1} \right)}^2}} \right]
\end{align}
where $\mathbf{\tilde x} = G(\mathbf{z},\mathbf{c}(y))$, $\mathbf{\widehat{x}} = \alpha \mathbf{x} + (1 - \alpha )\mathbf{\tilde x}$, and $\lambda $ is the gradient penalty weight.
The first and second terms approximate the Wasserstein distance, and the third term is the gradient penalty term for D. Unlike the condition GAN, the discriminator here is $D:\mathcal{X} \times \mathcal{E} \to \mathbb{R}$, which removes the sigmoid layer and outputs a real value. $\mathcal{L}_{\rm WGAN}$ cannot guarantee that the synthetic features are well suited for training a discriminative classifier. We conjecture that this issue could be alleviated by encouraging the generator to
construct features that have strong discriminating ability and can correctly reconstruct the word vector. To this end, following~\cite{felix2018multi}, we add a decoder to reconstruct the word vector of the synthetic feature. The cyc (cycle-consistency) loss function is used, which is given by
\begin{align}
{\mathcal{L}_{\rm CYC}} = {\left\| {\mathbf{\hat c}(y) - \mathbf{c}(y)} \right\|_2}
\end{align}
where $\mathbf{\hat c}(y)$ represents the reconstructed word vector. The final objective for training the feature generation network is as follows:
\begin{align}
{\min _G}~{\max _D}~ ({\mathcal{L}_{\rm WGAN}} + \beta {\mathcal{L}_{\rm CYC}})
\end{align}
where $\beta$ is hyper-parameter.

\subsection{Classifier training based on graph attention network}
To extract information from implicit representations (word vectors) and explicit representations (knowledge graphs), we use GCN to train classifier. Similar to~\cite{gao2019know}, we construct a knowledge graph whose nodes are word vectors of the concepts of seen classes, unseen action classes  and objects, and whose edges are defined by using the adjacency matrix later.

Let $S=|\mathcal{Y}^s|, U=|\mathcal{Y}^u|,  O = |\mathcal{O}|$.  GCN takes the word vectors of $ S + U$ classes and $O$ objects information as input, and obtains {all action class classifiers} $\{ {\mathbf{w}_i}\} _{i = 1}^{S+U}$ and all object classifiers $\{ {\mathbf{w}_i}\} _{i = S+U + 1}^{S+U + O}$ through the transmission and calculation of information between each layer of GCN. Among them, $O$ object classifiers act as a bridge between seen classes and unseen classes. Each layer of GCN takes the feature matrix of the previous layer $\mathbf{Z}^{(l - 1)}$ as input and outputs a new feature matrix $\mathbf{Z}^{(l)}$. The input of the first layer is a $k_{0} \times (S+U + O)$ feature matrix $\mathbf{Z}^{(0)}$, where $k_{0}$ represents the dimension of the input feature. The convolution operation of each layer in the network can be expressed as
\begin{align}
\mathbf{Z}^{(l)} = \mathbf{{D}}^{-\frac12} \mathbf{\widehat{A}} \mathbf{{D}}^{-\frac12} \left(\mathbf{Z}^{(l - 1)}\right)^{\top} \boldsymbol{\Phi}^{(l - 1)},~~l=1,\ldots,L
\end{align}
where $L$ denotes the total layer number,  
$\mathbf{Z}^{(l)} \in \mathbb{R}^{k_{l} \times (S+U + O)}$ represents the feature matrix of the $l$th layer, $1\leq l\leq L$,
$\mathbf{\widehat{A}} = \mathbf{{A}}+ \mathbf{I}$, $\mathbf{A}\in \mathbb{R}^{(S+U + O) \times (S+U + O)}$ represents the adjacency matrix of the knowledge graph, $\mathbf{{D}}\in \mathbb{R}^{(S+U + O) \times (S+U + O)}$ is a diagonal matrix whose diagonal entries are given by
${[\mathbf{{D}}]_{ii}} = \sum_j [\mathbf{\widehat{A}}]_{ij}$,  and $\boldsymbol{\Phi}^{(l - 1)}\in \mathbb{R}^{k_{l-1} \times k_{l}}$ represents the parameter matrix of the  $(l-1)$th layer. Each layer is followed by a ReLU function.

To further assign the trained action classifier with a stronger classification ability, we perform attention mechanism on the nodes, which update the relationship between action-object, object-object and action-action after each iteration. Specially, we define  the attentional coefficient matrix $\mathbf{B}$ as
\begin{equation}\label{G_RKHS}
[\mathbf{B}]_{ij}=
\begin{cases}
\frac{\mathbf{w}_{i}^{\top}\mathbf{w}_{j}}{\|\mathbf{w}_{i}\|_{2}\|\mathbf{w}_{j}\|_{2}},&\text{if $\mathbf{w}_{i}\in\mathcal{N}_k(\mathbf{w}_{j})$~or~$\mathbf{w}_{j}\in\mathcal{N}_k(\mathbf{w}_{i})$ }\\
0,&\text{otherwise}
\end{cases}
\end{equation}
where ${{\cal N}_k}({\mathbf{w}_j})$ represents the set of the $k$ nearest neighbors of the $j$th node. To make coefficients easily comparable across different nodes, we normalize them across all choices of  node $j$ using the softmax function by:
\begin{equation}\label{G_RKHS2}
[\mathbf{A}]_{ij}=\dfrac{\exp([\mathbf{B}]_{ij})}{\sum_{j' \in {{\cal N}(j)}}\exp([\mathbf{B}]_{ij'})}
\end{equation}
where ${{\cal N}(j)}$ represents the index set of all the neighbors of node $j$.

The cross entropy loss function of GCN in training is given by:
\begin{equation}\begin{split}
\mathcal{L}_{\rm CE} &=  - \frac{1}{N}\sum_{n = 1}^N {\sum_{c = 1}^{S+U} {y_n^c} } \log \left( {p_n^c} \right) \\
p_n^c  &= \frac{\exp \left( {\mathbf{w}_c^{\top} \mathbf{x}_n^c} \right)}{\sum_{c' = 1}^{S+U} \exp \left( \mathbf{w}_{c'}^{\top} \mathbf{x}_n^{c'} \right)}
\end{split}\end{equation}
where $y_n^c\in\{0,1\}$ is a boolean variable indicating whether the $n$th sample belongs to the $c$th class, $N$ represents the sum of the number of training samples of the seen class and the number of synthetic samples of the unseen class, $\mathbf{w}_c$ represents the classifier of $c$th action class, $\mathbf{x}_n^c$ represents the $n$th sample which belongs to the $c$th class, and $\mathbf{w}_c^{\top} \mathbf{x}_n^c$ represents the predicted score {of the $n$th sample in the $c$th class}, and $p_n^c$ represents the predicted score with a softmax operation.

\subsection{A two-stage deep network}
After training the feature generation network, we generate features of unseen classes by trained generator. Specifically, given the word vector $\mathbf{c}(u)$ of {an} unseen class $u \in {\mathcal{Y}^u}$ and random Gaussian noise $z \in {\cal Z}$, any feature $\mathbf{\tilde x}$ can be synthesized by $\mathbf{\tilde x} = G(z,\mathbf{c}(u))$, thus the synthesized training set $\widetilde {\cal U} = \{ (\mathbf{\tilde x},u,\mathbf{c}(u))\} $ can be obtained, which provides more training samples of unseen classes for the training classifier of graph convolution network based on attention mechanism, and enhances its classification performance and generalization ability. {Therefore, we propose a two-stage network, which joints feature generation network and graph attention network as shown in Figure~\ref{framework}, which is analyzed from both sample level and classifier level.} In this way, the training samples include the sample features of the seen class and unseen classes. During training, we first generate the features of unseen classes by sample level, and use them together with the features of seen classes as training samples. Then, both the classifiers of the seen classes and unseen classes are trained by classifier level. In the test stage, we use the classifier of unseen classes to classify test {videos}.

\section{Experiments}\label{Experiments}
\begin{figure*}[t]
\centering
\includegraphics[width=0.09\textwidth]{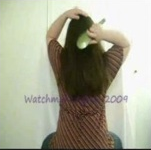}
\includegraphics[width=0.09\textwidth]{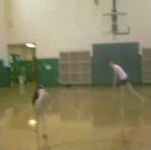}
\includegraphics[width=0.09\textwidth]{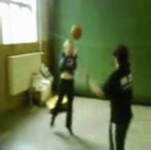}
\includegraphics[width=0.09\textwidth]{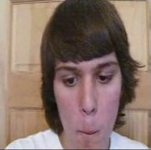}
\includegraphics[width=0.09\textwidth]{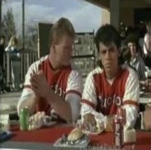}
\includegraphics[width=0.09\textwidth]{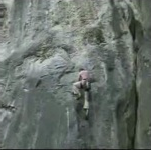}
\includegraphics[width=0.09\textwidth]{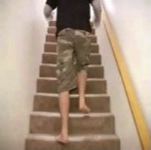}
\includegraphics[width=0.09\textwidth]{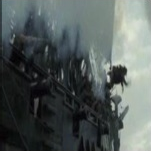}
\includegraphics[width=0.09\textwidth]{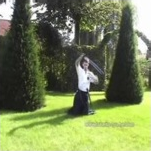}
\includegraphics[width=0.09\textwidth]{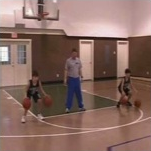}
\put(-505, -7){\small{brush~hair~~~cart~wheel~~~~~~catch~~~~~~~~~~chew~~~~~~~~~~
clap~~~~~~~~~~~climb~~~~~climb~stairs~~~~~~dive~~~~~~draw~sword~~~~~dribble}}\\
\includegraphics[width=0.09\textwidth]{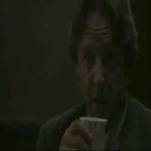}
\includegraphics[width=0.09\textwidth]{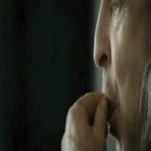}
\includegraphics[width=0.09\textwidth]{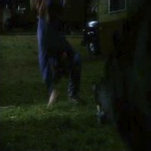}
\includegraphics[width=0.09\textwidth]{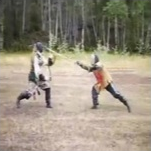}
\includegraphics[width=0.09\textwidth]{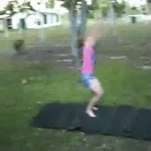}
\includegraphics[width=0.09\textwidth]{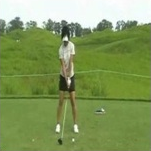}
\includegraphics[width=0.09\textwidth]{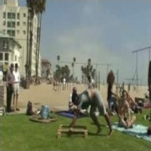}
\includegraphics[width=0.09\textwidth]{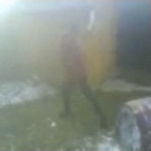}
\includegraphics[width=0.09\textwidth]{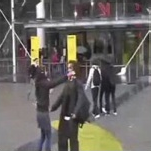}
\includegraphics[width=0.09\textwidth]{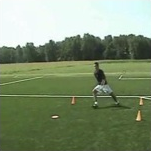}
\put(-497, -7){\small{drink~~~~~~~~~~~~eat~~~~~~~~~fall~floor~~~~~~fencing~~~~~~
flic~flac~~~~~~~~~~glof~~~~~~~hand~stand~~~~~~~~hit~~~~~~~~~~~hug~~~~~~~~~~~jump}}\\
\includegraphics[width=0.09\textwidth]{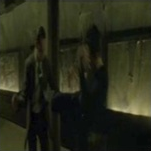}
\includegraphics[width=0.09\textwidth]{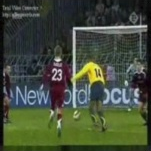}
\includegraphics[width=0.09\textwidth]{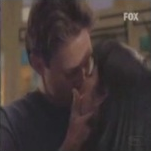}
\includegraphics[width=0.09\textwidth]{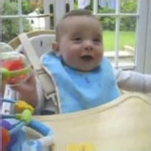}
\includegraphics[width=0.09\textwidth]{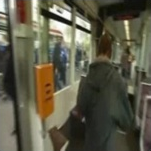}
\includegraphics[width=0.09\textwidth]{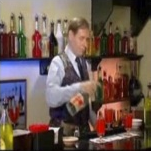}
\includegraphics[width=0.09\textwidth]{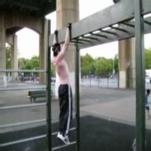}
\includegraphics[width=0.09\textwidth]{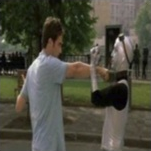}
\includegraphics[width=0.09\textwidth]{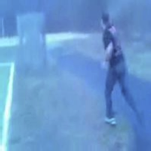}
\includegraphics[width=0.09\textwidth]{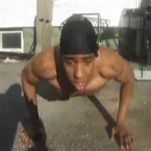}
\put(-495, -7){\small{kick~~~~~~~~~kick~ball~~~~~~~~~kiss~~~~~~~~~~~laugh~~~~~~~~~
pick~~~~~~~~~~~pour~~~~~~~~~pull~up~~~~~~~~punch~~~~~~~~~push~~~~~~~~~push~up}}\\
\includegraphics[width=0.09\textwidth]{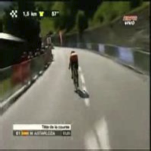}
\includegraphics[width=0.09\textwidth]{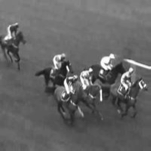}
\includegraphics[width=0.09\textwidth]{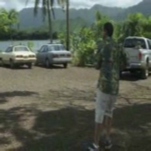}
\includegraphics[width=0.09\textwidth]{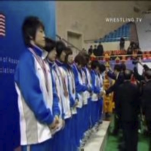}
\includegraphics[width=0.09\textwidth]{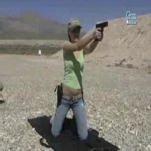}
\includegraphics[width=0.09\textwidth]{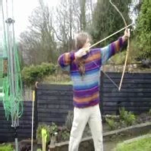}
\includegraphics[width=0.09\textwidth]{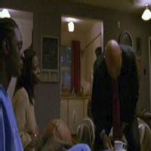}
\includegraphics[width=0.09\textwidth]{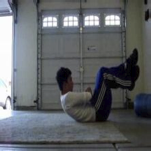}
\includegraphics[width=0.09\textwidth]{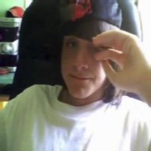}
\includegraphics[width=0.09\textwidth]{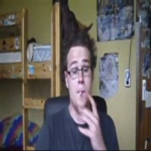}
\put(-502, -7){\small{ride~bike~~~~ride~horse~~~~~~~~run~~~~~~~shake~hands~~shoot~bow~~~shoot~gun~~~~~~~~sit~~~~~~~~~~~
sit~up~~~~~~~~~smile~~~~~~~~~~smoke}} \\
\includegraphics[width=0.09\textwidth]{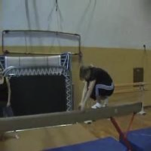}
\includegraphics[width=0.09\textwidth]{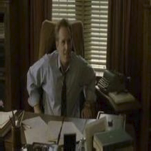}
\includegraphics[width=0.09\textwidth]{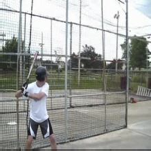}
\includegraphics[width=0.09\textwidth]{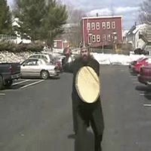}
\includegraphics[width=0.09\textwidth]{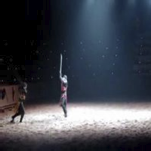}
\includegraphics[width=0.09\textwidth]{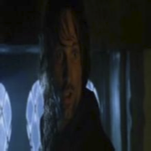}
\includegraphics[width=0.09\textwidth]{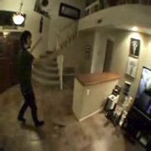}
\includegraphics[width=0.09\textwidth]{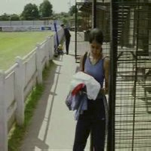}
\includegraphics[width=0.09\textwidth]{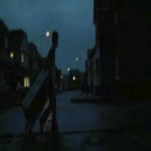}
\includegraphics[width=0.09\textwidth]{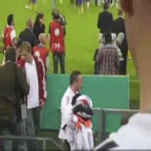}
\put(-507, -7){\small{somersault~~~~~~stand~~~swing~baseball~~~~sword~~~sword~exercise~~~~~talk~~~~~~~~~~throw~~~~~~~~~~
turn~~~~~~~~~~walk~~~~~~~~~~wave}} \\
\caption{{Some samples of  \texttt{HDMB51}, which consists of 51 action classes.}}
\label{hdmb51}
\end{figure*}

\begin{figure*}[t]
\centering
\includegraphics[scale=0.95]{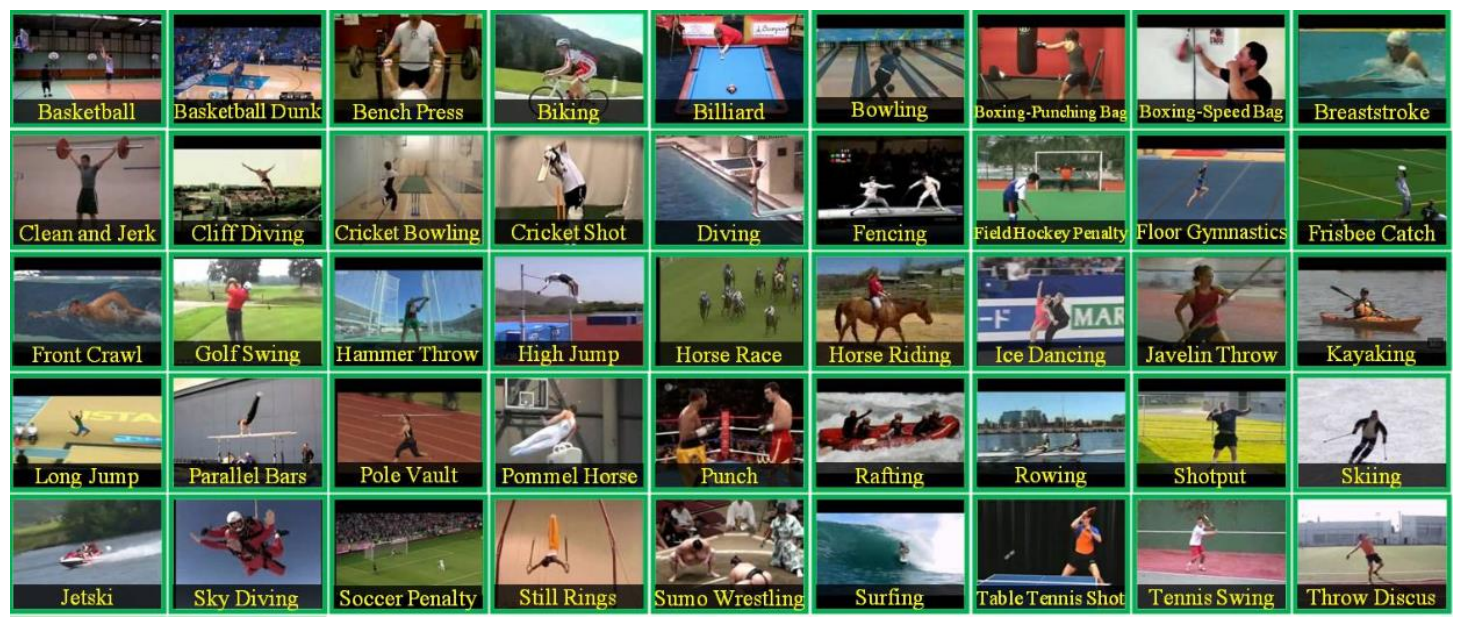}
\caption{{Some samples of  \texttt{UCF101}, which which consists of 101 action classes.}}
\label{ucf101}
\end{figure*}

This section verifies the effectiveness and superiority of FGGA on \texttt{UCF101}~\cite{soomro2012ucf101} and \texttt{HMDB51}~\cite{kuehne2011hmdb}. \texttt{HMDB51} consists of 6,766 videos, including 51 action classes, while  \texttt{UCF101} consists of 13,320 videos, including 101 action classes (see Figures~\ref{hdmb51}-\ref{ucf101}). We verify the effectiveness of FGGA using two different tasks: ZSL and GZSL. During the training phase, samples and attributes of seen classes are available for both ZSL and GZSL tasks. In the test stage, for ZSL, the trained model only evaluates data of unseen classes; For GZSL, the trained model evaluates the data of both seen and unseen classes.

\subsection{Experimental settings}
$\mathbf{Video~feature~extraction}$~~~~We use Inflated 3D~\cite{carreira2017quo} as the initial feature of video. Appearance Inflated 3D features and stream Inflated 3D features are extracted from mixed 5c layer output of RGB Inflated 3D network and stream Inflated 3D network, respectively. {Similar to~\cite{mandal2019out}, both networks are pre-trained on  \texttt{Kinetics} dataset~\cite{carreira2017quo}.}
These two features are 4096 dimensions respectively, and the feature of each video is the splicing of these two features, i.e., 8192 dimensions.
$\mathbf{Knowledge~graph~building}$~~~~We use ConceptNet~\cite{speer2017conceptnet} to build knowledge graph. ConceptNet is a semantic network that contains a lot of information about the world that computers should know to help them do better searches and understand human intentions. It consists of nodes representing concepts that are expressed as natural language words or phrases and whose relationships are indicated. ConceptNet contains relational knowledge and is connected to a subset of DBPedia that includes knowledge extracted from information box Wikipedia. Wikipedia is the largest collaborative online encyclopedia with 3 million articles in English. Much of this knowledge comes from Wiktionary, a free multilingual dictionary that provides information about synonyms, antonyms, and concepts translated into hundreds of languages. More dictionary knowledge comes from open multilingual WordNet, which is an English vocabulary containing more than 100,000 word concepts, rich in semantic information.
Something about intuitive associations of people for words comes from ``games with a purpose". Similar to~\cite{mettes2017spatial}, we use an English subgraph of about 1.5 million nodes and string matching to map concepts to nodes in ConceptNet. The most important thing in building a knowledge graph is to determine the relationships between these nodes. Specifically, if two nodes that connect to an edge can be found in ConceptNet, the initial correlation between the two nodes is expressed using the weight corresponding to the edge. Although the knowledge graph may have many types of edges, we follow the previous approach~\cite{mettes2017spatial} to reduce it to a single matrix to effectively express semantic consistency and propagate information between nodes.

$\mathbf{Word~embedding}$~~~~We use the images and videos from \texttt{YFCC100M}~\cite{mettes2017spatial} to train skip Gram network. The training model generates a 500-dimensional word vector representation for each word.

$\mathbf{Network~structure}$~~~~For our GAN, its generator is a three-layer fully connected network, whose output feature dimension is equal to video feature dimension, and the size of hidden layer is 4096. The decoder is also a three-layer fully connected network. The feature dimension of the output layer is equal to the word vector dimension of classes, and the size of the hidden layer is 4096. {The discriminator} is a two-layer fully connected network,  {of which the size of the dimension of the output/hidden  layer feature  is 1/4096},  {and the value of $\beta$} is 0.01. For our GCN, it is composed of three graph convolution layers, and the output channel sizes are 8192, 4096 and 4096 respectively. We {apply} LeakyReLU as the activation function after each convolutional layer of the graph and performed $\ell_2$ regularization on {generated classifiers}.

\subsection{Results}
In this section, we introduce the comparison results between FGGA and other ZSL methods on ZSL and GZSL tasks, and take the average accuracy and standard deviation as the evaluation criteria for effectiveness. First we compared the results on the ZSL task, then we compared the results on the GZSL task.
The comparative methods mainly include:
\begin{description}
\item[SVE]  (Self-training  with SVM and semantic Embedding~\cite{xu2015semantic})  performs self-training and data augmentation strategy
\item[ESZSL] (Embarrassingly Simple Zero Shot Learning~\cite{romera2015embarrassingly})  models the relationship between features and attributes
\item[SJE]  (Structured Joint Embedding~\cite{akata2015evaluation})  learns a compatibility function, which relates image features and side information
\item[MTE]  (Multi-Task Embedding~\cite{xu2016multi}) Performs multi-task visual-semantic mapping
\item[ZSECOC]  (Zero-Shot with Error-Correcting Output Codes~\cite{qin2017zero}) performs category-level error-correcting output codes
\item[GA]  (Generative Model~\cite{mishra2018generative}) models each action class as a probability distribution
\item[UR]  (Universal Representation~\cite{zhu2018towards}) performs generalized multiple-instance learning and universal representation learning
\item[TS-GCN]  (Two-Stream Graph Convolutional Network~\cite{gao2019know}) models the three types of relationships by two stream GCN
\item[ConSE]  (Convex combination of semantic embedding~\cite{norouzi2013zero}) performs image embedding system and semantic word embedding
\item[CLSWGAN]  (WGAN with class-level semantic information~\cite{xian2018feature}) generates features instead of images and is trained with a novel classification loss
\item[CEWGAN]  (WGAN with class-embeddings~\cite{mandal2019out})  performs WGAN
with cosine embedding and cycle-consistency to synthesize CNN features of unseen class
\item[CEWGAN-OD]  (WGAN with class-embeddings and OD detector~\cite{mandal2019out}) performs CEWGAN with out of distribution detector to synthesize CNN features of unseen class
\end{description}

\begin{table*}[!htb]
\caption{ZSL performance comparison with state-of-the-art methods on \texttt{HMDB51} and \texttt{UCF101}.}
\centering
\begin{tabular}{cccccc}
\hline
Method&	Reference&	Feature	& Label embedding&	\texttt{HMDB51}&	\texttt{UCF101}\\
\hline
SVE~\cite{xu2015semantic}&	ICIP2015&	Bag of words&	word2vec&	13.0$\pm$2.7	&10.9$\pm$1.5\\
ESZSL~\cite{romera2015embarrassingly}&	ICML2015&	Fisher vector&	word2vec&	18.5$\pm$2.0&	15.0$\pm$1.3\\
SJE~\cite{akata2015evaluation}	&CVPR2015&	Fisher vector&	word2vec	&13.3$\pm$2.4	&9.9$\pm$1.4\\
MTE~\cite{xu2016multi}&	ECCV2016&	Fisher vector&	word2vec	&19.7$\pm$1.6&	15.8$\pm$1.3\\
ZSECOC~\cite{qin2017zero}&	CVPR2017&	Fisher vector&	word2vec	&22.6$\pm$1.2&	15.1$\pm$1.7\\
GA~\cite{mishra2018generative}&	WACV2018&	Convolution 3D&	word2vec	&19.3$\pm$2.1	&17.3$\pm$1.1\\
UR~\cite{zhu2018towards}&	CVPR2018&	Fisher vector&	word2vec	&24.4$\pm$1.6	&17.5$\pm$1.6\\
CLSWGAN~\cite{xian2018feature}	&CVPR2018&	Inflated 3D	&word2vec	&29.1$\pm$3.8&	25.8$\pm$3.2\\
CEWGAN~\cite{mandal2019out}	&CVPR2019&	Inflated 3D&	word2vec	&30.2$\pm$2.7	&26.9$\pm$2.8\\
FGGA&	Ours&	Inflated 3D&	word2vec	&$\mathbf{31.2}\pm\mathbf{1.7}$	&$\mathbf{28.3}\pm\mathbf{1.8}$\\
\hline
\end{tabular}
\label{ZSL_results}
\end{table*}

\begin{table*}[!htb]
\caption{GZSL performance comparison with state-of-the-art methods on \texttt{HMDB51} and \texttt{UCF101}.}
\centering
\begin{tabular}{cccccc}
\hline
Method&	Reference&	Feature	& Label embedding&	\texttt{HMDB51}&	\texttt{UCF101}\\
\hline
SJE~\cite{akata2015evaluation}&		CVPR2015&		Fisher vector	&	word2vec&		10.5$\pm$2.4&		8.9$\pm$2.2\\
ConSE~\cite{norouzi2013zero}&		ICLR2014&		Fisher vector	&	word2vec&		15.4$\pm$2.8&		12.7$\pm$2.2\\
GA~\cite{mishra2018generative}&		WACV2018&		Convolution 3D&		word2vec&		20.1$\pm$2.1&		17.5$\pm$2.2\\
TS-GCN~\cite{gao2019know}&		AAAI2019&		object scores&		word2vec&		21.9$\pm$3.7&		33.4$\pm$3.4\\
CLSWGAN~\cite{xian2018feature}&		CVPR2018&		Inflated 3D&		word2vec&		32.7$\pm$3.4&		32.4$\pm$3.3\\
CEWGAN-OD~\cite{mandal2019out}&		CVPR2019&		Inflated 3D	&	word2vec	&	36.1$\pm$2.2&		37.3$\pm$2.1\\
FGGA&		Ours&		Inflated 3D&		word2vec&	$\mathbf{36.4}\pm\mathbf{2.0}$&		$\mathbf{37.6}\pm\mathbf{1.8}$\\
\hline
\end{tabular}
\label{GZSL_results}
\end{table*}

To make a fair comparison  on the ZSL task, we {use} the partitioning strategy in~\cite{xu2017transductive}, that is, 50\% of the action class videos are used for training, and the other 50\% of the action class videos are used for testing. For each dataset, we randomly {select} 10 partitions. Table~\ref{ZSL_results} shows experimental results on the ZSL task {with respect to different types of feature extraction: Bag of words,  Fisher Vector, Inflated 3D, with the same label embedding method: word2vec}. As can be seen from the table, FGGA achieves the best accuracy on \texttt{UCF101}  and \texttt{HMDB51}. Compared with traditional methods (ZSECOC, UR, etc.),
those methods based on deep learning (including FGGA) produce better performance. Compared with the recent methods UR, CLSWGAN and CEWGAN, FGGA improves {6.8\%/10.8\%, 2.1\%/2.5\% and 1\%/1.4\% respectively on \texttt{HMDB51}/\texttt{UCF101}}. These results demonstrate the effectiveness of FGGA. In addition, the performance of FGGA has a smaller standard deviation, which indicates that FGGA has a relatively stable identification performance under different training and testing data partition.

To make a fair comparison  on the GZSL task, we {use} the strategy in~\cite{mishra2018generative}, that is, 20\% of samples of the seen classes were selected for testing and the rest samples were trained. Table~\ref{GZSL_results} shows the experimental results on the GZSL task. It can be seen from the table that FGGA is superior to other comparative methods on the two datasets, reaching 36.4\% and 37.6\% recognition accuracies respectively on \texttt{HMDB51} and \texttt{UCF101}. Compared with {CEWGAN which achieved best results  currently}, FGGA achieves better performance.

\subsection{Effectiveness analysis}
\begin{figure}[t]
\centering
{\includegraphics[width=3in]{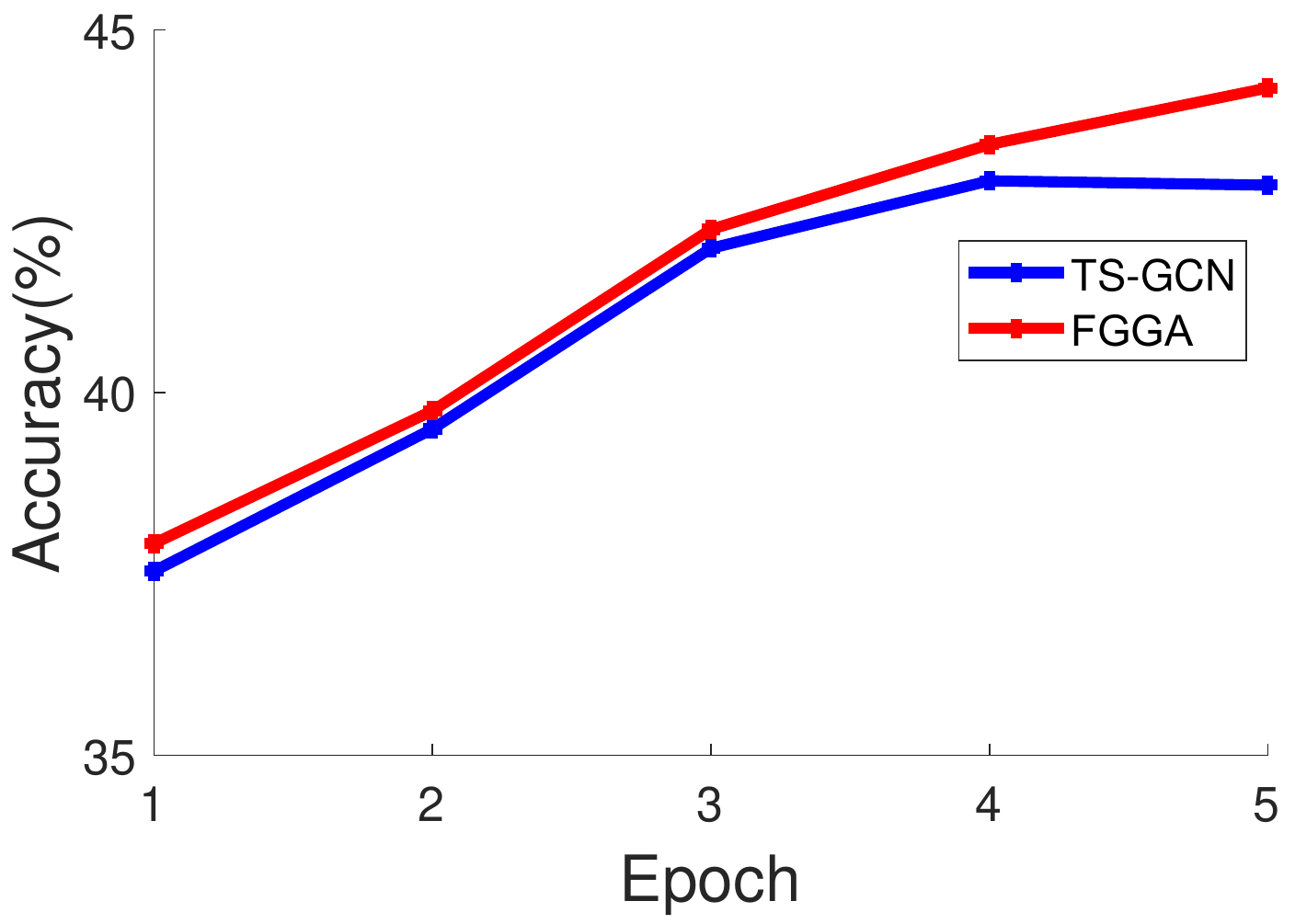}}
\caption{Comparison results of TS-GCN and FGGA with different epoches on  \texttt{UCF101}.}
\label{TSGCN}
\end{figure}

In this section, we further compare with several baseline methods. We first compare the performance of FGGA and TS-GCN. To make a fair comparison with TS-GCN, we also use object scores feature~\cite{gao2019know} as the initial feature, and represent each sample  by a graph structure. Therefore, the feature generation network is not applicable. Figure~\ref{TSGCN} shows the comparison results of FGGA and TS-GCN under different epochs on \texttt{UCF101}, mainly to verify the effectiveness of the graph attention mechanism. As can be seen from the figure, the results of the two methods are equivalent when $epoch\leq 3$. when $epoch>3$ the result of TS-GCN tends to be stable, FGGA still has an upward trend. In general, FGGA has better classification performance, which proves that the graph attention mechanism is effective.

Table~\ref{baseline_results} shows quantitative results of  FGGA versus several GAN based ZSL methods, which includes the accuracy of the seen class, the accuracy of the unseen class, and the harmonic mean of them.
Among these comparative methods, CLSWGAN applied GAN to ZSL earlier, and CEWGAN-OD is the best ZSL method based on GAN at present. As can be seen from Table~\ref{baseline_results}, FGGA achieves the best accuracy. The results of FGGA and CEWGAN-OD were significantly better than those of CLSWGAN and CEWGAN. Compared with the two methods, FGGA improved the performance  {by 3.7\%/5.2\% and 2.8\%/3.9\% respectively on \texttt{HMDB51}/\texttt{UCF101}}, which indicated that the design of a reasonable classifier could significantly improve the performance. Compared with CEWGAN-OD, FGGA is slightly less accurate in {unseen classes}, but it is significantly better in {seen classes}. The reason for better accuracy performance of CEWGAN-OD on unseen classes may be that CEWGAN-OD adds an OD detector which can detect whether each sample belongs to a seen class before classification, thus improving the classification performance. However, FGGA {merely uses a common classifier for classification instead of OD detectors}, which still achieves considerable performance.
{To analyze the sensitivity of FGGA to different word embedding methods, and make a fair comparison with CEWGAN, we test the performance on another Word2Vec embedding trained on GoogleNews as shown in Table~\ref{GoogleNews_results}. From the table we can see that although the performance of FGGA has some changes after changing the word vector, it can still achieve better performance.}
{Overall}, the above analysis demonstrates that FGGA outperforms the baseline methods.
\begin{table}[!htbp]
\centering
\caption{Accuracy of GZSL versus state-of-the-art methods on \texttt{HMDB51} and \texttt{UCF101}.}\label{baseline_results}
\renewcommand{\multirowsetup}{\centering}
{
\begin{tabular}{>{\centering\arraybackslash}ccccccc}
\hline
\multirow{2}{*}{Methods}&	\multicolumn{3}{c}{\texttt{HMDB51}}	& \multicolumn{3}{c}{\texttt{UCF101}}\\
	&Seen class	&Unseen class	&Harmonic mean	&Seen class	&Unseen class	&Harmonic mean\\
\hline
CLSWGAN	&52.6	&23.7&	32.7&	74.8	&20.7&	32.4\\
CEWGAN&	51.7	&24.9	&33.6	&73.7	&21.8&	33.7\\
CEWGAN-OD&	55.6	&26.8	&36.1&	75.9	&24.8&	37.3\\
FGGA&	57.5&	26.6	&$\mathbf{36.4}$&	78.3&	24.7	&$\mathbf{37.6}$\\
\hline
\end{tabular}}
\end{table}

\begin{table*}[!htb]
\caption{{ZSL performance comparison with word vector trained on GoogleNews.}}
\centering
\begin{tabular}{cccccc}
\hline
Method&	\texttt{HMDB51}&	\texttt{UCF101}\\
\hline
CEWGAN~\cite{mandal2019out}& 30.2	&26.9\\
FGGA&		$\mathbf{31.5}$	&$\mathbf{27.8}$\\
\hline
\end{tabular}
\label{GoogleNews_results}
\end{table*}

\begin{figure}[t]
\centering
{\includegraphics[width=3in]{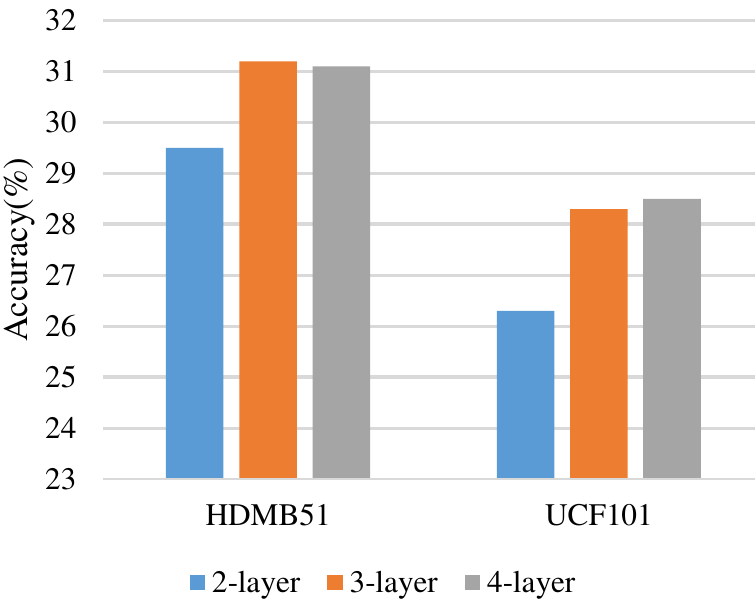}}
\caption{Comparison results with different number of GCN layers on  \texttt{HMDB51} and \texttt{UCF101}.}
\label{GCNcanshu}
\end{figure}

\begin{figure}[t]
\centering
{\includegraphics[width=2.5in]{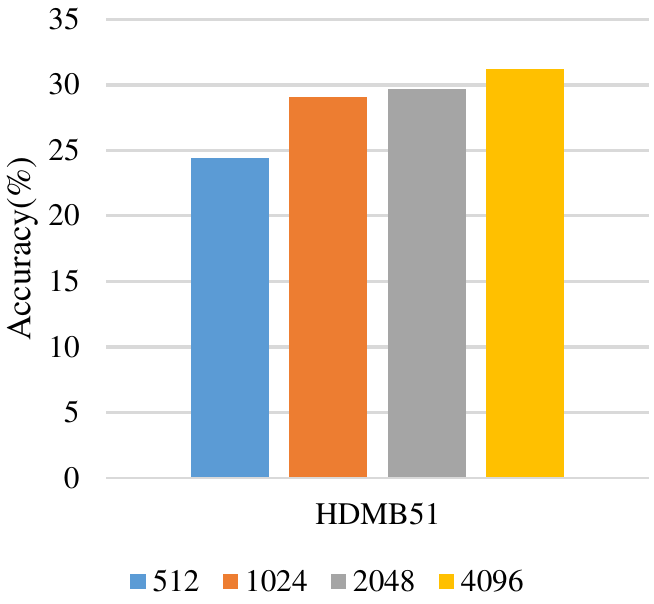}}
\caption{Comparison results of different feature dimensions on  \texttt{HMDB51}.}
\label{GAN_feature}
\end{figure}

\subsection{Parameter analysis}
To verify the effect of GCN depth on FGGA performance, we compared FGGA performance at different levels. Figure~\ref{GCNcanshu} shows the results of FGGA under different GCN layers, in which the number of output channels of two layers is 4096,4096 respectively. For the three layers, the number of output channels is 8192, 4096, 4096 respectively. The number of output channels for the three layers is 8192, 4096, 4096, 4096 respectively. As can be seen from the figure, the performance of the three-layer GCN is the same as that of the four-layer GCN, and the number of deeper layers does not get a higher recognition rate. The reason may be that the number of training samples is not sufficiently large, leading to an overfitting of the deeper network.

To verify the influence of feature dimensions on FGGA, we compared the performance of FGGA under different feature dimensions, as shown in Figure~\ref{GAN_feature}. As can be seen from the figure, when the feature dimension is 512, the performance is obviously poor. When the feature dimension reaches 1024 dimensions, the effect is significantly improved, and the best recognition accuracy can be obtained when the feature dimension reaches 4096 dimensions.

\subsection{Ablation study}

\begin{figure}[t]
\centering
{\includegraphics[width=3in]{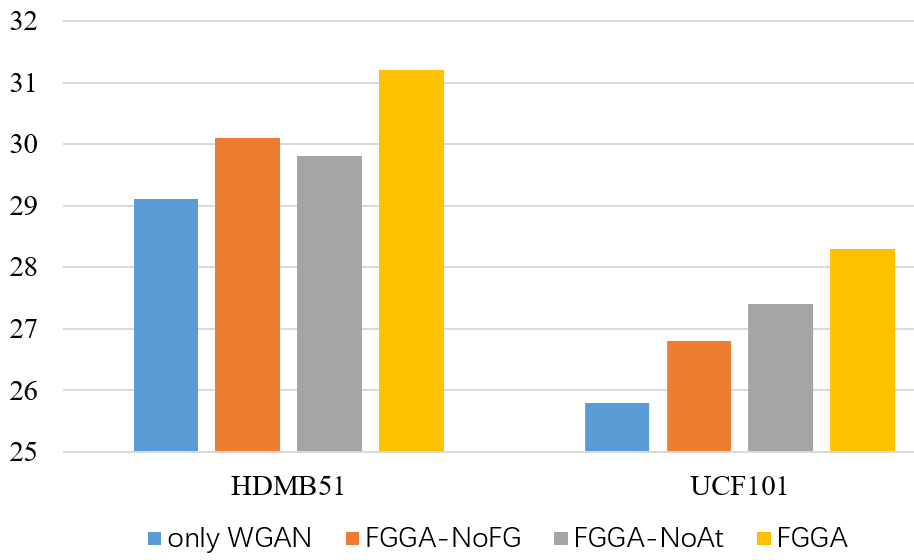}}
\caption{Ablation results of FGGA on  \texttt{HMDB51} and \texttt{UCF101}.}
\label{ablation}
\end{figure}

To further verify the effectiveness of {feature generation network and graph attention network, we conducted ablation experiments by comparing FGGA with FGGA without using feature generation network (denoted by FGGA-NoFG), FGGA without using attentional mechanism (denoted by FGGA-NoAt) and model with only WGAN (denoted by only WGAN).  Figure~\ref{ablation} shows the comparative results, which indicates that feature generation network and graph attention convolutional network introduced in FGGA are both important and effective for ZSL.} As for FGGA-NoFG, this method has no synthesis features of unseen classes in training, which {leads to a poor classification performance of the learning classifier for unseen classes}. For FGGA-NoAt, this method does not update the adjacent matrix dynamically in training, which prevents the relationship among action-action, action-object and object-object from being adjusted adaptively, so the classifier classification performance is poor. {For only WGAN, this method has no cycle-consistency constraint term, and doesn't analyse the relationship of different classes. It results in the worst performance.}
Therefore, each component of FGGA {contributes to FGGA in an indispensable fashion}.
\section{Conclusion}
\label{Conclusion}
In this paper, we propose a new zero-shot action recognition framework: FGGA, which can effectively improve generalization and discrimination of model by analyzing relationship between the action classes and training sample imbalances of seen and unseen class. On the one hand, FGGA effectively circumvents  the problem of unbalanced training samples of seen classes and unseen classes by establishing and generating the action features of the unseen class generated by the conditional Wasserstein GAN with additional loss terms. On the other hand, FGGA transfers the knowledge learned from the seen class to the unseen class by means of the prior knowledge graph, and proposes the graph convolution network based on the attention mechanism, which leads to promising classification performance of learned classifiers. Experimental results show that FGGA can achieve higher accuracy on two public datasets.
Comprehensive performance studies have been conducted by comparing FGGA with state-of-the-art methods over two datasets. The effectiveness of FGGA is evidenced by its favorable performances compared with others.

\end{document}